\documentclass[letterpaper, 10 pt, conference]{ieeeconf}  

\IEEEoverridecommandlockouts                              

\usepackage{amsmath,amsfonts}
\usepackage{algorithmic}
\usepackage{algorithm}
\usepackage{array}
\usepackage[caption=false,font=normalsize,labelfont=sf,textfont=sf]{subfig}
\usepackage{textcomp}
\usepackage{stfloats}
\usepackage{url}
\usepackage{verbatim}
\usepackage{graphicx}
\usepackage{cite}
\usepackage{hyperref}
\usepackage{setspace}
\usepackage{tabularx}
\usepackage{pifont} 
\usepackage{booktabs}

\usepackage{subcaption} 
\usepackage[font=footnotesize]{caption}
\title{\LARGE \bf
Bridging Semantics and Kinematics: A Modular Framework for Zero-Shot Robotic Manipulation}

\usepackage{tcolorbox}
\usepackage{courier}

\author{Ali Alabbas$^{*1}$, Dipshikha Das$^{*1}$, Camillo Murgia$^{*1}$, Sainul Ansary$^{2}$, Alaa Elkamash$^{2}$, Philip Long$^{2}$ 
\thanks{*Authors have equally contributed to this work}%
\thanks{$^{1,2}$Authors are with the Department of Electronic, Software and Advanced Manufacturing Engineering,
        Atlantic Technological University, Galway, Ireland
        {$^{1}$\tt\small (ali.alabbas, dipshikha.das, camillo.murgia)@research.atu.ie}
        {$^{2}$\tt\small (sainul.ansary, alaa.elkamash, philip.long )@atu.ie}}%
}

\begin{document}

\maketitle
\thispagestyle{empty}
\pagestyle{empty}

\begin{abstract}

This paper presents a modular training-free framework for zero-shot, language-guided robotic manipulation in semi-structured environments. The architecture bridges the gap between high-level reasoning and low-level kinematics by decomposing the vision-action pipeline into three stages: visual perception, semantic interpretation, and task execution. To overcome the spatial ambiguity and semantic hallucinations inherent in standard Vision-Language Models (VLMs), the perception module employs FastSAM and Set-of-Mark (SoM) prompting to dynamically generate grounded, alphanumeric visual anchors. The same foundation model then operates purely as a Large Language Model (LLM) to act as a semantic router, translating unconstrained human directives into verifiable, reconfigurable configurations. Finally, these configurations are dynamically parsed by a Task Orchestrator into MoveIt Task Constructor (MTC) to generate collision-free trajectories. The framework is evaluated across two zero-shot experimental setups: unconstrained open-world sequential manipulation and dense relational spatial reasoning, achieving a $62\%$ end-to-end task success rate across both scenarios, demonstrating its capacity to reliably execute complex physical actions without domain-specific training or manual coordinate programming. https://youtu.be/UIYFkwCB9nA

\end{abstract}

\section{INTRODUCTION}

Humans are able to collaborate efficiently in a shared workspace to accomplish common tasks, supported by the ability to interpret nonverbal cues and predict each other's intentions. Replicating such seamless shared autonomy between a human and a robot, however, remains challenging. In this paper, we present a novel framework in which a robot interprets minimal natural language commands from a human operator and autonomously executes the intended manipulation task. The main objectives include making the system modular and training-free by leveraging LLMs and VLMs as core reasoning components. Consequently, the system adapts to new tasks without any domain-specific fine-tuning. The framework is organized into three tightly coupled components: visual perception, semantic interpretation, and kinematic task execution. 

Perception is an integral part of the pipeline, focusing particularly on zero-shot understanding of objects and spatial relationships within a scene. The rise of vision foundation models~\cite{han2025multimodal} has shifted the field from traditional deep learning-based methods toward the use of off-the-shelf VLMs, driven by the need for broader generalisation across object categories and environments. While learning-based detection~\cite{redmon2016you} and instance segmentation~\cite{zhao2023fast,cheng2022masked} methods perform well on a defined set of object classes or scenarios, they often struggle to generalise to unseen objects or novel scenarios. Open-vocabulary object detection methods such as Grounding DINO~\cite{liu2024grounding}, OwlViT~\cite{minderer2022simple}, and GLIP~\cite{li2022grounded} aim to address this limitation, yet they frequently struggle with visual aliasing in dense environments.
Nevertheless, vision models come with their own set of limitations~\cite{hu2025large}, such as difficulty distinguishing between visually identical items within a scene, hallucinations, weak spatial reasoning, and sensitivity to prompts. One approach to mitigate some of these limitations is careful prompt design, but this often leads to long and coordinate-heavy prompts, which contradicts the goal of enabling minimal, natural human commands.

\begin{figure}[t]
    \centering
    \includegraphics[width=\columnwidth]{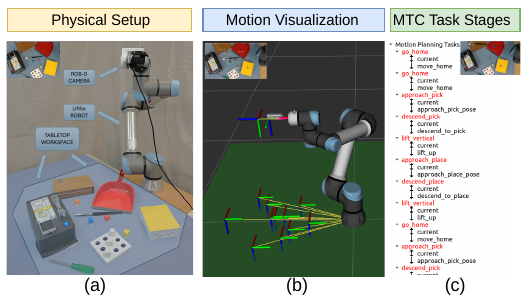}
    {\footnotesize
    \caption{Hardware platform and system overview: (a) \textit{Physical Setup:} The UR5e equipped with RGB-D wrist camera operating over a tabletop workspace containing diverse general objects and the current observation view, (b) \textit{Motion Visualization:} The segmented objects and their corresponding published frames over the observed image, (c) \textit{MTC Task Stages:} The compositional stage hierarchy automatically generated from a single natural command, decomposed into parameterized MTC primitives. }
    \label{fig:robot}
    }
\vspace{-14pt}    
\end{figure}

While VLMs demonstrate strong high-level scene understanding, they struggle to translate into precise, actionable outputs for performing physical manipulation tasks \cite{gao2024physically}. Recent trends have attempted to address this through Vision-Language-Action (VLA) models~\cite{din2507vision,kawaharazuka2025vision}, particularly robotic foundation models~\cite{zitkovich2023rt,team2510gemini,zhou2025chatvla,openvla,saycan} that adopt an end-to-end approach to map raw pixels and text directly to joint actions. While powerful, these architectures present deployment bottlenecks. They typically require massive parameter counts, extensive domain-specific fine-tuning, and heavy compute resources, creating a bottleneck when adapting the system to novel tasks or environments. Consequently, they are difficult to deploy locally on standard robotic hardware. Moreover, their end-to-end nature introduces significant interpretability and safety challenges. These limitations have motivated a system design that integrates off-the-shelf VLMs into a structured, multi-stage architecture, combining unified visual grounding with declarative task execution, enabling interpretable and computationally efficient zero-shot manipulation across reconfigurable task primitives.

\begin{figure*}[t]
    \centering
    \includegraphics[width=\textwidth]{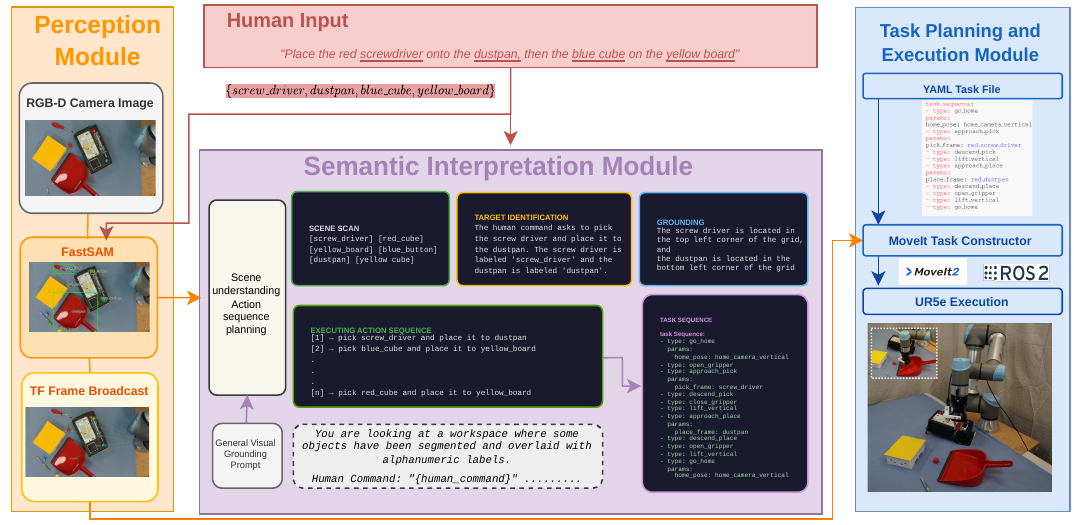}
    \caption{System architecture overview illustrates the end-to-end flow from free-form natural language to coordinated multi-step robot motion. The pipeline comprises three modules: \textbf{(left) Perception Module}, which processes RGB-D camera input through FastSAM segmentation and broadcasts object transformation frames (tf) to establish a shared spatial vocabulary; \textbf{(centre) Semantic Interpretation Module}, which receives the natural language command and annotated scene representation, performs a structured reasoning chain via Molmo2, and outputs a validated YAML task sequence; and \textbf{(right) Task Planning and Execution Module} where the Task Orchestrator ingests the YAML configuration, compiles it into executable MoveIt Task Constructor pipelines within the ROS 2 framework, and drives physical execution on the UR5e manipulator.}
    \label{fig:pipeline}
\vspace{-14pt}      
\end{figure*}

Classical task and motion planning (TAMP) approaches, such as the Planning Domain Definition Language (PDDL)~\cite{aeronautiques1998pddl} and symbolic planners like GraphPlan~\cite{blum1997fast} and FastForward~\cite{hoffmann2001ff}, require manual task modelling, and are limited to structured environments. To overcome these limitations, there is increasing interest in using LLMs for their flexibility in complex, dynamic environments~\cite{chen2024autotamp, ding2023integrating}. These foundation models offer an intuitive interface for generating high-level task plans~\cite{singh2022progprompt} and action sequences~\cite{wang2024llm, huang2022language} from natural language descriptions without task-specific fine-tuning. However, the quality of these generated plans relies heavily on prompt design. Prior work has optimized performance using techniques such as ontological knowledge~\cite{din2026onto}, fixed template-based prompts~\cite{wang2024llm}, and interactive prompting~\cite{li2023interactive}.

\begin{table}[h]
\renewcommand{\arraystretch}{1.2} 
\setlength{\tabcolsep}{2pt} 

\begin{tabularx}{\columnwidth}{@{} l *{5}{>{\centering\arraybackslash}X} @{}}
\toprule
\textbf{Paper} & \textbf{Gener-} & \textbf{Inter-} & \textbf{Zero-} & \textbf{Modu-} & \textbf{Param.} \\
\textbf{Ref.} & \textbf{alisation} & \textbf{pretability} & \textbf{shot} & \textbf{lar} & \textbf{Binding} \\
\midrule
\cite{zitkovich2023rt}   & \ding{51} &     &   \ding{51}      &           & \\
\cite{team2510gemini}    & \ding{51} &  &  \ding{51}         &           & \\
\cite{openvla}           & \ding{51} &  &   \ding{51}      &           & \\
\cite{zhou2025chatvla}   & \ding{51} &  &     \ding{51}  &           & \\
\cite{saycan}            & \ding{51} & \ding{51} & \ding{51} & \ding{51} & \\
\midrule
Ours                     & \ding{51} & \ding{51} & \ding{51} & \ding{51} & \ding{51} \\
\bottomrule
\end{tabularx}
\caption{Summary of key features in related work on robotic manipulation frameworks.}
\label{tab:summary}
\vspace{-14pt} 
\end{table}

Enabling a robot to execute diverse manipulation tasks from natural language commands autonomously requires solving three tightly coupled problems: grounding visual scenes into task-relevant representations, translating unconstrained human intent into safe executable plans, and reliably executing those plans through contact-rich motions on physical hardware. Existing approaches (Table~\ref{tab:summary}) often sacrifice interpretability~\cite{cohen2024survey} by relying on end-to-end policy learning with limited explicit parameter binding, i.e., the automatic assignment of scene-specific values such as object identifiers, poses, and coordinates to abstract slots in motion primitives and modular task reasoning ~\cite{kwon2025kinodynamic}.
We propose a modular framework that addresses these challenges through a declarative task architecture, representing manipulation tasks as structured, human-readable configurations. This separation between what a task is and how it is executed enables zero-shot adaptation to novel instructions without modifying the underlying motion planning stack. This work contributes an advanced systems engineering framework that integrates off-the-shelf foundation models to achieve zero-shot, training-free manipulation. Specifically, our contributions are:

\begin{itemize}
    \item An attention-driven visual grounding pipeline for scene understanding by combining FastSAM-based~\cite{zhao2023fast} instance segmentation with Set-of-Mark visual prompting. 
    \item A reconfigurable task structure built from a library of modular task primitives, orchestrated by a constrained LLM to translate high-level human commands.
    \item A YAML-driven task architecture for zero-shot manipulation adaptation that is instantiated at runtime from scene-specific values, and executed using MTC~\cite{gorner2019moveit}.
\end{itemize}

The remainder of the paper is organized as follows: Section ~\ref{Sec: Methods} details the proposed system architecture. Section ~\ref{Sec: Exp} presents the experimental setups and end-to-end system evaluation. Finally, Section ~\ref{Sec: Conclusion} concludes the paper and discusses limitations and future research directions.

\section{METHODOLOGY}  \label{Sec: Methods}
We present a training-free framework for zero-shot autonomous manipulation that integrates an open-weight foundation model (VLM/LLM) with a declarative motion planning architecture in ROS 2. As illustrated in Fig.~\ref{fig:pipeline}, the system operates across three stages: (1) visual scene understanding via FastSAM and SoM grounding, (2) semantic command translation to instantiate task parameters, and (3) motion planning and execution via YAML-driven task orchestration and MoveIt Task Constructor (MTC)~\cite{gorner2019moveit} on a UR5e manipulator. The following subsections describe each stage in detail.

\subsection{Multi-modal Perception Module}
To achieve zero-shot execution of natural language commands, the perception module must reliably interpret unstructured human intent within visually complex workspaces. A primary challenge in such environments is object disambiguation, where workspaces often contain multiple geometrically and visually identical items such as uniform screws, identical fasteners, or multiple identical assembly slots. While powerful foundation models continue to advance in grounding and spatial reasoning, their reliance on cloud infrastructure and massive compute introduces costs and latency, rendering them impractical for real-time robotic deployment~\cite{cheng2024spatialrgpt}. Conversely, efficient open-weight models are ideal for local execution, but they inherently struggle to differentiate these visually identical items without dense, coordinate-heavy prompting~\cite{google2025gemini3provision}.

To resolve this, our system employs an attention-driven, Set-of-Mark visual prompting strategy~\cite{yang2023set}. Given the task context, the system isolates the relevant Region of Interest (ROI), then applies FastSAM~\cite{zhao2023fast} to segment the scene and dynamically overlay unique identifiers onto task-relevant elements. This augmented visual feedback establishes a shared spatial vocabulary, enabling the VLM to comprehend the scene state without requiring exhaustive prompt engineering.

This architecture closely mirrors human-human collaborative workflows. Just as a human co-worker can interpret and execute a concise, high-level directive such as "\textit{insert the remaining screws into the top bracket}" by relying on shared visual context, our system leverages these segmented identifiers to immediately ground the user's intent. By integrating the user's minimal command with the annotated visual state, the system accurately enumerates targeted objects and publishes their transformations to ROS2 tf tree.

More importantly, dedicated spatial reasoning in the perception layer drastically reduces the cognitive burden on the foundation model. This eliminates the need for massive, cloud-based architectures and instead enables the use of efficient, open-weight VLMs (8B Molmo2 \cite{clark2026molmo2}), deployed entirely locally. By simply reading explicit visual identifiers, the VLM can achieve highly accurate relational reasoning.

\subsection{Semantic Interpretation Module}
Once the visual workspace is semantically grounded, the system translates the human’s free-form directive into executable robotic behaviors. Direct end-to-end translation approaches where models map natural language directly to joint trajectories or raw executable code often suffer from physical unpredictability and logic hallucinations, posing significant safety risks in human-robot collaborative environments~\cite{bhat2024grounding}.

To address this, our architecture utilizes the local Molmo2 model as a cognitive intermediary to map human intent to a deterministic intermediate Domain-Specific Language (DSL). The model takes as input both the user’s natural language instruction and the structured scene representation produced by the perception module. We also provide a general visual grounding prompt which supports the generation of a desired action sequence, as shown in Figure~\ref{fig:pipeline}. Operating in a zero-shot setting, the model functions as a semantic router that aligns the human’s objective with a logical sequence of predefined, modular task primitives such as $pick-and-place$ or $button\_press$. It then instantiates these primitives by assigning the required spatial parameters using the previously defined visual identifiers (e.g., mapping $cube$ to the $pick\_frame$ and $board$ to the $place\_frame$). The model is constrained to populate only the task sequence of a pre-validated YAML schema. This configuration file serves as the definitive bridge between the semantic reasoning and kinematic execution pipeline. If the semantic router generates an unfeasible sequence (e.g., wrong picking frame), the kinematic planner fails to find a valid motion plan.

\subsection{Task Planning and Execution Module}

\subsubsection{YAML-Driven Task Generalization}
The validated YAML configuration generated by the semantic interpretation module is processed by the execution module, which performs action parameter binding and translates the symbolic task descriptors into an MTC pipeline, effectively acting as a dynamic compiler between high-level task representations and physical robot motions. The YAML schema is structured with two layers: a \textit{task template library} defining parameterized motion patterns, and a \textit{task instance} definition produced by the LLM that instantiates the templates with scene-specific values. The task templates encode reusable motion strategies such as pick-and-place and constrained push, using defined stage types, planner configuration, and constraints. At runtime, a task orchestration layer resolves these definitions by substituting values from the LLM-generated task instance, such as specific target coordinates (\textit{place\_frame: "point\_1"}) or action constraints (\textit{press: "red\_button"}). This process autonomously binds the abstract symbolic intent to concrete physical targets without requiring any manual code updates.

\subsubsection{MoveIt Task Constructor Integration}

Once the task hierarchy is fully parameterized, motion stages in the task sequence are interpreted by a stateless Task Interpreter that translates each YAML stage specification into the corresponding MTC stage object. The interpreter ensures that task plans are reproducible and that stage construction is deterministic given identical input configurations. MTC provides a compositional framework for motion planning in which complex manipulation tasks are expressed as a set of planning stages with explicit data flow between them, as shown in Fig. ~\ref{fig:robot}(c). The Cartesian and joint-space planners, along with robust inverse kinematics (IK) solvers, generate smooth collision-free trajectories, and the planner parameters are specified per stage from the YAML, decoupling the high-level definition from the low-level kinematic implementation. 

The task orchestration binds the overall structure by resolving the template variables from the current scene state, assigning motion stages to the MTC interpreter, and executing gripper and I/O stages. This layered architecture allows for a clean separation between \textit{task definition, task interpretation}, and \textit{task execution}. The changes to task logic require only YAML updates and are independent of the task template library and other layers. This enables the system to adapt to novel, zero-shot user instructions on the fly, transforming natural language and visual context into safe, precise physical actions without requiring any manual code reconfiguration from the human co-worker.

\begin{figure}[t]
    \centering
    \includegraphics[width=\columnwidth]{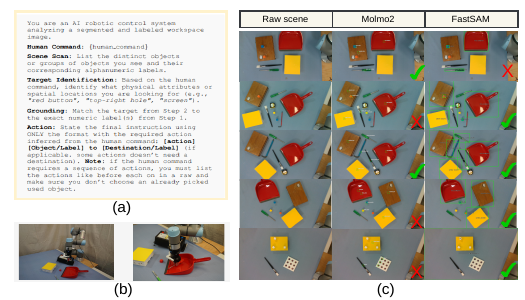}
    {\footnotesize
    \caption{(a) \textit{General visual grounding prompt} template for visual grounding and action generation illustrating the reasoning chain, (b) \textit{UR5e execution} during open-world manipulation trials, (c) \textit{Ablation study} comparing object grounding performance across scene configurations \textit{(rows)} using the raw scene baseline, Molmo2 alone, and FastSAM with SoM prompting. Green ticks indicate successful object grounding and correct identifier assignment; red crosses indicate grounding failures.}
    \label{fig:results}
    }
\vspace{-14pt}      
\end{figure}

\section{EXPERIMENTAL RESULTS} \label{Sec: Exp}

The hardware setup consists of a UR5e robotic arm, a Robotiq Hand-E gripper, and an Intel RealSense D435 depth camera, mounted on the robot's wrist, as shown in Figure~\ref{fig:robot}. The system was executed locally on a dedicated workstation equipped with a single NVIDIA RTX A4500 GPU (20GB VRAM) and 64 GB of RAM. The software stack runs within the ROS2 framework, with MTC providing the motion planning interface and the ROS2 driver of the UR5e handling the low-level joint commands.

Because this framework focuses on zero-shot semantic routing rather than low-level control optimization, we evaluate task-level performance using three coarse, sequential success metrics for all trials:
\begin{itemize}
\item \textbf{Perception Success:} The system successfully localizes, segments, and assigns visual identifiers to all task-relevant objects, and they match the active ROS transformation frames (\textit{tf}).
\item \textbf{Translation Success:} The VLM generates a structurally valid, chronologically accurate YAML task sequence strictly utilizing the defined MTC physical primitives.
\item \textbf{Execution Success:} MTC successfully parses the configuration, calculates valid, collision-free kinematics, and executes the physical trajectory.
\end{itemize}

\subsection{Open-World Sequential Manipulation} 
This experiment evaluates the system’s capacity for zero-shot, language-guided manipulation in a semi-structured, cluttered environment. It tests the framework's ability to generalize to unseen objects and formulate multi-step orchestration without prior task configuration.

\subsubsection{Experimental Setup and Procedure}
The workspace contains a diverse mixture of general objects (e.g., boxes, screwdrivers, task board) placed randomly, as shown in Fig~\ref{fig:robot}. A human operator issues unconstrained, multi-step commands by displaying text on a mobile device within the robot's camera field of view (e.g., \textit{"Start from home, then press the red button, then pick the stylus and place it on the white board, then return home."}). As a preliminary proof-of-concept for our system, we conducted $30$ trials, varying both the target objects and their spatial configurations in the workspace.

\subsubsection{Results}
Table~\ref{tab:main_results} shows the results across the $30$ evaluation trials, where the integrated pipeline achieved a $70\%$ end-to-end task success rate, successfully orchestrating and completing the unconstrained multi-step commands. To isolate the specific failure modalities of the foundation model, we evaluate the system across three sequential conditional gates: perception, translation, and execution. The system achieved a $76.7\% $ Perception Success rate. The failures stemmed from two distinct sources. First, FastSAM occasionally made part-to-whole segmentation errors, such as misidentifying a mallet's handle as a 'pencil' or a single leg of red pliers as a 'red pen.' Second, the VLM (Molmo2) struggled with fine-grained size comparisons. While it successfully identified the 'biggest' or 'smallest' dustpan when multiple dustpans were present, it failed to reliably isolate the 'biggest' cube among a group of similar cubes. Finally, in one trial, the VLM failed to recognize that a dustpan was already occupied, indicating the inability to infer functional states purely from visual appearance.



Of the $23$ trials that passed the perception stage, the system achieved a $91.3\%$ Translation Success rate. The translation failures were mainly due to strong linguistic priors, where the VLM associated the 'press' action incorrectly with button-like affordances. This led to a hallucinated frame name and triggered an illogical command: \textit{"press the paint brush."}, producing an invalid YAML sequence. 


Finally, for the $21$ trials that produced valid YAML sequences, the system achieved a $100\%$ Execution Success rate. This proves that once the VLM generates a correctly formatted and physically logical semantic plan, the MTC pipeline can execute the zero-shot manipulation reliably.


\begin{table}[htbp]
\caption{System Evaluation Success Rates}
\label{tab:main_results}
\centering
\resizebox{\columnwidth}{!}{%
\begin{tabular}{|l|c||c|c|c|}
\hline
\textbf{Task Scenario} & \textbf{End-to-End} & \textbf{Perception} & \textbf{Translation} & \textbf{Execution} \\
\hline
Open-World Manipulation & \textbf{70\% (21/30)} & 76.7\% (23/30) & 91.3\% (21/23) & 100\% (21/21) \\
Dense Spatial Reasoning & \textbf{53.3\% (16/30)} & 53.3\% (16/30) & 100\% (16/16) & 100\% (16/16) \\
\hline
\textbf{Overall (Average)} & \textbf{62\% (37/60)} & 65.0\% (39/60) & 94.8\% (37/39) & 100\% (37/37) \\
\hline
\end{tabular}%
}
\vspace{-14pt}  
\end{table}

\subsection{Spatial and Relational Instruction Mapping}
This experiment evaluates the system's capacity for zero-shot spatial reasoning. Specifically, it assesses how effectively the system can ground abstract, relational, or under-specified human commands such as \textit{"fill the top row of holes with pegs"} strictly using dynamically generated visual labels, eliminating the need for hard-coded Cartesian coordinates or task-specific prompt tuning.

\subsubsection{Experimental Setup and Procedure}
The workspace featured unstructured objects alongside dense arrays of visually identical targets, specifically $3\times3$, $5\times5$ and circular pegboards. The human operator provided concise, natural commands such as \textit{"fill the upper half of the holes with the available pegs"} or \textit{"fill the empty holes on the board"} without providing any coordinate information. We conducted $30$ trials varying the starting state of the board and the requested spatial relations.

\subsubsection{Results}
Across the $30$ evaluation trials, the system achieved a $53.3\%$ end-to-end task success rate. As shown in Table~\ref{tab:main_results}, the perception and grounding successfully parsed the scene and generated the correct sequence of targeted interactions in $53.3\%$ of the cases. Notably, the system demonstrated zero-shot state estimation during non-intuitive commands such as \textit{"fill the holes adjacent to the filled one."} Without explicit state-tracking memory, the VLM successfully analyzed the Set-of-Mark image to differentiate between occupied and unoccupied holes, to generate a task sequence for picking random pegs and placing them into the target holes.

The failures occurred when the VLM struggled with precise relational mapping within dense target arrays. In the $5\times5$ pegboard trials, the most frequent failure mode was spatial offset in row identification, such as when instructed to \textit{"fill the middle row,"} the system selected targets in the second row rather than the third.  In other cases, the model hallucinated a single corner position as the target region, including commands targeting the diagonal placements, representing spatial failure under ambiguous relational commands. In the case of the circular board with 12 holes where the configuration lacked spatial patterns, the task success degraded significantly as the system implicitly relied on the structural regularity to perform relation grounding tasks such as \textit{"fill the holes clockwise"}, where the unstructured layout affected the grounding between image and reasoning. 


For the $16$ trials that successfully passed the spatial grounding phase, the system achieved a $100\%$ Translation Success and a $100\%$ Execution Success rate. The translation module formatted the VLM's output into valid YAML sequences, and the Task Orchestrator, along with MTC, successfully translated and executed the physical placement of all pegs.

\subsection{Ablation Study: Impact of Set-of-Mark Grounding}
To quantify the necessity of the dedicated perception module, we compared our full architecture against a baseline where unannotated images were fed directly to the Molmo2 VLM, while also being fed with human commands. The only difference is that the visual grounding prompt (Fig. ~\ref{fig:results}) does not mention the labeled objects. We evaluated 10 baseline and 10 proposed trials, evenly split between the open-world and dense spatial setups.

In the open-world environment, the baseline achieved a $60\%$ success rate, outputting imprecise or shifted target coordinates, and once labeling the wrong object. The proposed pipeline achieved $80\%$, with its sole failure caused by linguistic semantic ambiguity (confusing a "blue cube" with a "wooden box") rather than spatial inaccuracy.

The limitation of the baseline is most severe in the dense pegboard task, which lacks explicit numeric anchors, and the VLM succumbs to severe visual aliasing among identical items, resulting in a $0\%$ success rate. By offloading coordinate extraction to the FastSAM and SoM layer, our proposed pipeline clearly identified the scene, achieving $100\%$ success for task execution. This contrast proves that our visual grounding layer is essential for enabling efficient, open-weight VLMs to perform high-precision spatial reasoning.

\section{Conclusion and Future Work} \label{Sec: Conclusion}
We presented a modular, training-free framework for zero-shot, language-guided robotic manipulation that integrates an off-the-shelf VLM with a declarative, YAML-driven execution architecture via MoveIt Task Constructor. By decomposing the vision-to-action pipeline into visual grounding, semantic interpretation, and kinematic execution, the system achieves zero-shot adaptation to novel instructions without retraining or manual code reconfiguration. The SoM visual grounding mechanism effectively resolves the spatial ambiguity and visual aliasing limitations of standard VLMs. Experimental evaluations across open-world manipulation and dense spatial reasoning tasks demonstrate the system's capacity to reliably translate unconstrained human directives into precise robotic actions, achieving a $62\%$ overall end-to-end task success rate.

Despite these capabilities, the architecture presents notable limitations. A primary challenge is the foundation model's strong linguistic priors, which can occasionally override explicit visual data, leading to hallucinated action frames. Additionally, while FastSAM accurately segments most objects, it occasionally struggles with part-to-whole segmentation and geometrically similar items. Future iterations could integrate a stronger open-vocabulary model, such as Grounding DINO, to improve spatial perception.

Future work will explore constrained decoding and active prompting to enforce stricter adherence to generated visual labels. To overcome the limitations of single-viewpoint 2D segmentation in dense, occluded environments, we plan to investigate 3D semantic point clouds and multi-view segmentation for complex 6-DoF manipulation. Furthermore, to address the static nature of the current kinematic primitives, future iterations will focus on closing the execution loop, enabling the system to detect failures and dynamically replan using real-time hardware feedback. Finally, we aim to conduct comparative benchmarking against end-to-end Vision-Language-Action (VLA) models to further quantify the computational and interpretability advantages of our modular approach.

\addtolength{\textheight}{-1cm}   

\end{document}